\documentclass[11pt]{article}
\usepackage{acl}
\usepackage{graphicx}
\usepackage{booktabs}
\usepackage{multirow}
\usepackage{amsmath}
\usepackage{makecell}
\usepackage{enumitem}

\usepackage{tikz}
\usetikzlibrary{arrows.meta, positioning, fit, calc}

\usepackage[table]{xcolor}

\title{UMCR: Universal Medical Concept Recognition through Guideline Integration and Agentic Linking}
\author{Fabrice Lamarche \and Amal Zouaq \\
        LAMA-WeST Lab, Polytechnique Montr\'eal, Mila -- Quebec AI Institute}

\begin{document}

\maketitle

\begin{abstract}
Medical Concept Recognition is the task of extracting and grounding spans in medical text to entries in a standard ontology. Current methods are generally developed towards a single ontology and source text type (clinical notes, drug labels, etc.) The central challenge is therefore to build methods that adapt to any target ontology without requiring ontology-specific retraining.
We propose UMCR\footnote{Code will be provided upon acceptance. 
}, a pipeline that allows extraction and linking to any target ontology using any source text variety, utilizing direct guideline integration and an agentic linking module.
We evaluate our method on the recent MedPath benchmark \cite{mishra2025medpath}, covering a wide variety of medical text types and vocabularies, and show that our method achieves strong performance across these tasks. This ability unlocks interoperability in the healthcare domain, allowing for extraction and alignment between different medical standards. Our method also reduces dependence on parametric LLM memory, thanks to improved interaction with the target ontology compared to other proposed approaches.
\end{abstract}

\section{Introduction}
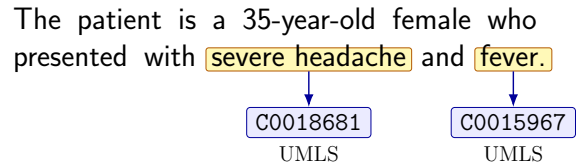
\begin{figure}[t]
    \centering
    \resizebox{\columnwidth}{!}{%
    \begin{tikzpicture}[
    font=\sffamily\LARGE,
    word/.style={
        text height=2ex,
        text depth=0.4ex
    },
    concept/.style={
        word,
        fill=yellow!35,
        draw=orange!70!black,
        rounded corners=1.5pt,
        inner xsep=3pt,
        inner ysep=0.8pt
    },
    umls/.style={
        draw=blue!60!black,
        fill=blue!8,
        rounded corners=2pt,
        inner xsep=5pt,
        inner ysep=4pt,
        font=\Large\ttfamily
    },
    arrow/.style={
        -{Latex[length=2.5mm]},
        thick,
        blue!60!black
    }
]

\node[word] (w1) {The};
\node[word, right=0.12em of w1] (w2) {patient};
\node[word, right=0.12em of w2] (w3) {is};
\node[word, right=0.12em of w3] (w4) {a};
\node[word, right=0.12em of w4] (w5) {35-year-old};
\node[word, right=0.12em of w5] (w6) {female};
\node[word, right=0.12em of w6] (w7) {who};

\node[word, below=0.15em of w1.south west, anchor=north west] (w8) {presented};
\node[word, right=0.12em of w8] (w9) {with};
\node[concept, right=0.12em of w9] (c1) {severe headache};
\node[word, right=0.12em of c1] (w10) {and};
\node[concept, right=0.12em of w10] (c2) {fever.};

\node[umls, below=7mm of c1] (u1) {C0018681};
\node[umls, below=7mm of c2] (u2) {C0015967};

\draw[arrow] (c1.south) -- (u1.north);
\draw[arrow] (c2.south) -- (u2.north);

\node[font=\large, below=1pt of u1] {UMLS};
\node[font=\large, below=1pt of u2] {UMLS};

\end{tikzpicture}
    }
    \caption{Concept Recognition Example.}
    \label{fig:cr_ex}
\end{figure}

Clinical notes, biomedical literature, social media posts about healthcare, and other document types all contain different information, which can be encoded using various medical vocabularies. Examples of such vocabularies include the Human Phenotype Ontology (HPO) \cite{gargano2024human}, covering phenotypic abnormalities, and RxNorm \cite{liu2005rxnorm}, which contains standardized nomenclature for clinical drugs. Automatic extraction of such data elements from free-form text is important as it allows for development of downstream applications like summarization, question-answering systems or outcome analysis \cite{labbe2023chatgpt}. \mbox{Figure}~\ref{fig:cr_ex} shows an example of the Concept Recognition task on a simple sentence, with linking to the UMLS ontology \cite{bodenreider2004unified}. Medical Concept Recognition is also crucial in interoperability, where it allows information from distinct Electronic Health Record (EHR) systems to be shared or compared \cite{brailer2005interoperability}, greatly improving the ease of use, deployment, and usefulness of these systems.

Designing a method that can work across medical text modalities is difficult because they all use different vocabularies and jargon, have a variable amount of information to encode, and do not follow the same format \cite{liu2024uncovering}. Furthermore, datasets and benchmarks make choices about exactly what counts as information to capture, and these decisions are typically captured in annotation guidelines given to human annotators during dataset curation  \cite{dougan2014ncbi, roberts2017overview}. Making direct use of these guidelines can allow for methods that adapt to varied tasks \cite{DBLP:conf/iclr/SainzGALRA24, shinohara2025efficient}, but this has not yet been studied in Medical Concept Recognition.

We address these issues directly; UMCR conditions the stages of its pipeline on a per-dataset compiled guideline and links spans against the Unified Medical Language System (UMLS) \cite{bodenreider2004unified}, a large collection of medical terms covering most sub-domains. UMLS contains cross-vocabulary mappings, allowing initial extractions to be directly converted to target vocabularies. Dataset nuances are captured in the compiled guidelines and allow our method to adapt to the specificity of each individual task. We use the term \textit{universal} in this sense: a single method allows extraction towards any target ontology, from varied text source types, without requiring any training data. Our main contributions are the following: 

\begin{itemize}[itemsep=2pt]
    \item A method for Medical Concept Recognition using Compiled Guidelines and Agentic Linking, allowing for extraction across ontologies and text modalities.
    \item Evidence of the universality of this method through evaluation over the nine MedPath datasets \cite{mishra2025medpath} and phenotype datasets BioC-GS \cite{weissenbacher2023phenoid} and GSC+\cite{gsc+}.
    \item Ablations isolating the impact of the different pipeline stages.
\end{itemize}

\section{Related Work}

\paragraph{Dictionary and rule systems.}
Initial Medical Concept Recognition approaches are based on pattern-matching between the target ontology and the input text. Examples include MetaMap \cite{Aronson2001MetaMap}, the NCBO Annotator \cite{Jonquet2009OBA}, Doc2HPO \cite{10_1093_nar_gkz386}, or QuickUMLS \cite{soldaini2016quickumls}. These systems are fast and precise, but suffer from two main limitations. First, while these systems might make use of ontology synonyms for given concepts, they still lack the flexibility to handle broader lexical diversity. Second, these systems are largely developed with a single ontology in mind, and fail to generalize when changing this vocabulary target (see Section \ref{sec:results}).

\paragraph{Supervised neural systems.}
To overcome these limitations, supervised approaches have been proposed, aiming to capture more nuance in the extracted concepts. 
PhenoTagger \cite{Luo_2021} builds a distantly supervised dataset from the target ontology and trains BERT \cite{DBLP:conf/naacl/DevlinCLT19} model variants on it. While this setup is ontology-dependent because of dedicated training, it can theoretically be adapted to work across target vocabularies, with mixed performance. PhenoGPT \cite{Yang2023PhenoBCBERT} is another example of a supervised system where a generative model is fine-tuned on target datasets, which improves accuracy but is limited to this fully supervised scenario.

\paragraph{LLM-based systems.}
Recent advances in Language Modeling have motivated direct application of Large Language Models (LLM) to the Concept Recognition task. REAL \cite{shlyk2024real} separates the task into mention detection and entity linking, prompting GPT-3.5 \cite{OpenAI2023GPT35} for extraction, and grounding using retrieval-augmented generation (RAG)\cite{lewis2020retrievalaugmented}. AutoPCR \cite{tao2025autopcr} combines hybrid rule-based and neural extractions using a pretrained BERT model for mention detection and only uses the LLM for the selection of the best matching candidate. 
Their shared limitation is heavy reliance on the language model's parametric memory, which poorly covers diverse ontologies. Evaluations of generative models for phenotype recognition \cite{groza2024evaluation} confirm that prompting LLMs without ontology information is unreliable for the linking part of the task, supporting RAG-inspired pipelines. \citet{larsen2026extraction} also shows that augmenting an LLM with tools for phenotype extraction improves performance across models but does not study the end-to-end recognition task.

\begin{figure*}[h]
    \centering
    \resizebox{\textwidth}{!}{%

\providecommand{\umcrgraphicon}{\tikz[baseline=-0.5ex, line width=0.5pt]{
    \draw (0,0) -- (0.25,0.18) -- (0.5,0);
    \draw (0.25,0.18) -- (0.25,-0.18);
    \fill (0,0) circle (1pt);     \fill (0.25,0.18) circle (1pt);
    \fill (0.5,0) circle (1pt);   \fill (0.25,-0.18) circle (1pt);}}
\providecommand{\umcrvecicon}{\tikz[baseline=-0.5ex, line width=0.5pt]{
    \draw (0,0.18) -- (0,-0.18) -- (0.5,-0.18);
    \fill (0.10,-0.07) circle (1pt); \fill (0.20,0.02)  circle (1pt);
    \fill (0.32,-0.05) circle (1pt); \fill (0.28,0.11)  circle (1pt);
    \fill (0.42,0.05)  circle (1pt);}}

\begingroup\sffamily
\begin{tikzpicture}[
    font=\small,
    >=Stealth,
    stage/.style={
        rectangle, rounded corners=4pt, draw, thick,
        minimum width=5.8cm, align=center, inner sep=3pt
    },
    chip/.style={
        rectangle, rounded corners=3pt, draw, thick,
        align=center, inner sep=3pt, font=\footnotesize
    },
    heading/.style={align=center, font=\footnotesize},
    container/.style={
        rectangle, rounded corners=12pt, draw=black, thick,
        dashed, inner sep=0.25cm
    },
    group/.style={
        rectangle, rounded corners=6pt, draw=black!60, thick,
        dashed, inner xsep=0.2cm, inner ysep=0.12cm
    },
    caption/.style={font=\small\bfseries, inner sep=0pt},
    flow/.style={-{Stealth[length=2.5mm]}, black, thick},
    feed/.style={-{Stealth[length=2.5mm]}, black!60, thick},
    link/.style={black, thick}
]


\node[heading] (ontology) at (0,0) {Ontology};

\node[stage, below=0.25cm of ontology,
    draw=black!60, fill=black!6] (stage0)
    {Stage 0: \textbf{Ontology profiling}};

\node[chip, anchor=north west, draw=black!60, fill=white]
    (graphtools) at ([shift={(0.2cm,-0.42cm)}]stage0.south west)
    {\umcrgraphicon \\[3pt] Graph tools};
\node[chip, anchor=north east, draw=black!60, fill=white]
    (embeds) at ([shift={(-0.2cm,-0.42cm)}]stage0.south east)
    {\umcrvecicon \\[3pt] Concept Embeddings};
\node[group, fit=(graphtools)(embeds)] (outpair) {};

\node[heading, below=0.3cm of outpair] (guideline)
    {Raw Guidelines};

\node[stage, below=0.25cm of guideline,
    draw=black!60, fill=black!6] (stage1)
    {Stage 1: \textbf{Guideline compilation}};

\node[chip, below=0.25cm of stage1, minimum width=5.8cm,
    draw=black!60, fill=white]
    (brief) {Compiled Guidelines (JSON)};

\draw[flow] (ontology.south)  -- (stage0.north);
\draw[flow] (stage0.south)    -- (outpair.north);
\draw[flow] (guideline.south) -- (stage1.north);
\draw[flow] (stage1.south)    -- (brief.north);

\node[container, fit=(ontology)(stage0)(outpair)(guideline)(stage1)(brief)] (offlinebox) {};


\node[stage, anchor=north west,
    draw=black!60, fill=black!3]
    (document) at ([shift={(1.4cm,-0.25cm)}]offlinebox.north east)
    {\textbf{Input} \\
    {\footnotesize $\ldots$ with
    severe headache and fever.}};

\node[stage, below=0.22cm of document,
    draw=blue!30!black!45, fill=blue!4] (stage2)
    {Stage 2: \textbf{Mention detection} \\
    {\footnotesize ``severe headache'', ``fever''}};

\node[stage, below=0.22cm of stage2,
    draw=green!30!black!45, fill=green!5] (stage3)
    {Stage 3: \textbf{Agentic linking} \\
    \footnotesize search ``headache'' $\to$ C0018681 (0.93) \\
    \footnotesize get\_children $\to$ C0017693 ...};

\node[chip, right=0.4cm of stage3,
    draw=black!60, fill=white, align=left] (toollist)
    {search\\
    get\_concept\\
    get\_parents\\
    get\_children\\
    get\_siblings\\
    adjust\_span\\
    commit\\
    unlink};

\node[stage, below=0.22cm of stage3,
    draw=purple!30!black!45, fill=purple!4] (stage4)
    {Stage 4: \textbf{Filtering} \\
    {\footnotesize ("headache", C0018681) = \texttt{KEEP} ...}};

\node[stage, below=0.22cm of stage4,
    draw=black!60, fill=black!3] (output)
    {\textbf{Output} (mention--concept pairs) \\
    {\footnotesize "headache" $\to$ C0018681} ...};

\node[container, fit=(document)(stage2)(stage3)(stage4)(output)(toollist)] (perdocbox) {};


\path let \p1=(offlinebox.south), \p2=(perdocbox.south)
    in coordinate (capY) at (0,{min(\y1,\y2)});
\node[caption, anchor=north] at ($(offlinebox.south |- capY)-(0,0.05cm)$)
    {Offline (once per dataset)};
\node[caption, anchor=north] at ($(perdocbox.south |- capY)-(0,0.05cm)$)
    {Per document};


\draw[flow] (document.south) -- (stage2.north);
\draw[flow] (stage2.south)   -- (stage3.north);
\draw[flow] (stage3.south)   -- (stage4.north);
\draw[flow] (stage4.south)   -- (output.north);
\draw[link] (stage3.east)    -- (toollist.west);   


\path let \p1=(offlinebox.east), \p2=(perdocbox.west)
    in coordinate (gapx) at ({(\x1+\x2)/2},0);
\coordinate (guideBus) at ($(gapx)-(0.25cm,0)$);
\coordinate (toolBus)  at ($(gapx)+(0.25cm,0)$);

\draw[feed] (brief.east) -- (guideBus |- brief.east) |- (stage2.west);
\draw[feed] (brief.east) -- (guideBus |- brief.east) |- ([yshift=3mm]stage3.west);
\draw[feed] (brief.east) -- (guideBus |- brief.east) |- (stage4.west);

\draw[feed] (outpair.east) -- (toolBus |- outpair.east) |- ([yshift=-3mm]stage3.west);

\end{tikzpicture}
\endgroup
    }
    \caption{UMCR pipeline.}
    \label{fig:pipeline}
\end{figure*}
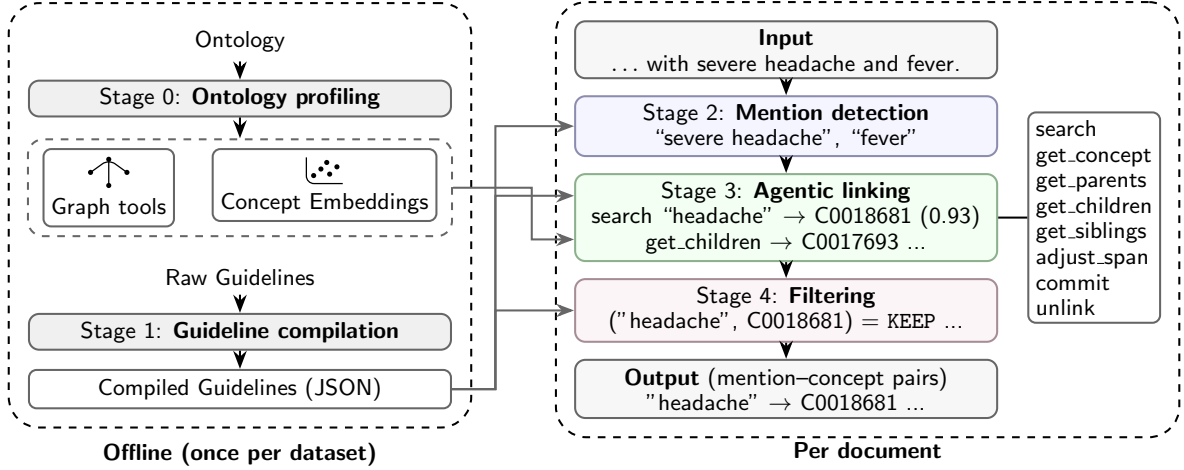

\paragraph{Evaluation.}
Many benchmarks exist for this task, each with specific target ontologies and source texts. Examples include the NCBI Disease corpus \cite{dougan2014ncbi}, which contains disease mentions from PubMed abstracts, linking to the Medical Subject Headings (MeSH) ontology, and the COMETA corpus \cite{basaldella2020cometa}, which annotates social media posts with SNOMED CT concepts. LLM-based work mainly reports performance on GSC+ \cite{gsc+} and BioC-GS from BioCreative VIII Track 3 \cite{weissenbacher2023phenoid}, which both use HPO as the target ontology.
The recently proposed MedPath \cite{mishra2025medpath} dataset regroups nine expert-annotated datasets into a single benchmark. This allows evaluation of methods across ontologies and source texts. See Section \ref{sec:datasets} for details on the dataset and medical vocabularies used in MedPath. 


\section{Methodology}
\label{sec:method}

Our method can be divided into offline pre-computation done once per ontology, and per-document processing to produce mention-identifier pairs from input text. Figure~\ref{fig:pipeline} illustrates this pipeline, and we describe each stage in detail next.

\paragraph{Stage 0 - Ontology Profiling}
This stage corresponds to the encoding of the ontology concepts, for later use in candidate retrieval. Here, each concept name is concatenated with its synonyms, and embedded using the lightweight Qwen3-Embedding-0.6B model \cite{qwen3embedding}\footnote{See Appendix \ref{app:impl} for details.}. This stage also includes the creation of our ontology graph exploration tools (\textit{graph tools}). These represent the code interface between the ontologies and the model, which allow the Agentic Linking step to explore the ontology concepts to find the best matches (Further details are given in the Stage 3 description). In practice, since we use subsets over UMLS to represent the different ontologies, we only define these tools once, but they could be created to adapt to any ontology that does not contain UMLS links.

\paragraph{Stage 1 - Guideline Compilation}
Guidelines for each dataset are typically distributed either through external documents or directly as text in the dataset material, code or paper, and can range from very detailed descriptions, with examples and rules, to short instruction sentences\footnote{See Figure \ref{fig:guideline-examples} in Appendix \ref{app:guideline_compilation} for examples}. To standardize this, we manually encode each guideline into a unified JSON format, which we call \textit{compiled guidelines}. This JSON format contains a short task summary, target mention categories with definitions, ontology constraints through rules and linking policy. An example of this guideline compilation step is presented in Appendix \ref{app:guideline_compilation}. This brings uniformity in the way in which dataset specificity is presented to the model during extraction and linking. We study how this compilation step impacts performance in Section \ref{sec:experiments}.

\paragraph{Stage 2 - Mention Detection}
In this stage, we prompt an LLM to extract text spans and use the compiled guidelines from Stage 1 for detailed instructions. The exact prompt for this step is shown in Appendix \ref{app:prompts}. We also augment the resulting mentions with exact matches between the source text and ontology concept names or synonyms.

\paragraph{Stage 3 - Agentic Linking}
Here, the mentions from Stage 2 are passed individually to the LLM, along with the compiled guidelines,  graph tools from the Ontology Profiling step (Stage 0) and the source text. The LLM then autonomously explores the ontology to find the most relevant concept for the given mention and guidelines. The available tools are listed in Figure \ref{fig:pipeline} and detailed in Appendix \ref{app:stage3-tools}. The model returns one JSON action per turn, with a maximum of 8 actions; \texttt{commit} and \texttt{unlink} are terminal. Ontology exploration is encouraged through prompting, and most linked mentions use multiple tool calls to refine the concept choice, with an average of 5.3 tool calls per mention. Tool usage is detailed further in Appendix \ref{app:tool_analysis}. The Agentic Linking prompt is shown in Appendix \ref{app:prompts}. Exact match mentions from Mention Detection (Stage 2) do not pass through agentic linking and are directly linked to the corresponding concept and passed to the Filtering step (Stage 4).

\paragraph{Stage 4 - Filtering}
This last step serves as a final check for guideline-following. Each mention-concept pair is passed individually to a LLM, tasked with assessing whether the pair respects the dataset guidelines, and returning a reject motive when a pair is discarded. The prompt used for Stage 4 is shown in Appendix \ref{app:prompts}. The impact of this step on the pipeline is detailed in Section \ref{sec:experiments}.

\section{Experiments}
\label{sec:experiments}
\subsection{Task and Metrics}
As different names are used to refer to Medical Concept Recognition across literature, we define the task as follows: Given an input text, a guideline specification, and a target ontology, a system produces a list of mention-identifier pairs. We evaluate systems on the Mention Detection step, where we use traditional NER measures of either Exact Match (EM) or Span Overlap (SO) \cite{segura-bedmar-etal-2013-semeval} to count valid extractions. We can then measure precision, recall, and F1 Score with respect to the gold mentions. Furthermore, we can evaluate the correctness of the linking step by counting a matched candidate mention as correct only if its identifier is exactly the same as the gold annotation. This is the main metric for which we report our results, as it evaluates the ability of a model to identify concepts and link them correctly.

\subsection{Datasets}
\label{sec:datasets}
To evaluate our approach against other methods, we use the nine datasets from the MedPath benchmark, as well as the two common phenotype-based datasets for Medical Concept Recognition, BioC-GS and GSC+. This covers a diverse set of input text types (social media posts, paper abstracts, etc.) as well as domain, length, target ontology and guideline specificity.
Table~\ref{tab:datasets} details the eleven datasets used. Systems were run on the respective test sets of the datasets when available. CADEC, MIMIC-IV-EL and Mantra-GSC were not provided with specific splits so we use the full corpora. We use the training split for the ShAReCLEF dataset because the test annotations are not publicly available.

\begin{table*}[h]
\centering
\resizebox{\textwidth}{!}{
\begin{tabular}{@{}llrrl@{}}
\toprule
Name (abbrev.) & Sents & Domain & Target ontology \\
\midrule
ShAReCLEF (SC) \cite{suominen2013overview} & 7,957 & Hospital discharge notes & SNOMED CT \\
Mantra-GSC (MAN) \cite{kors2015mantra} & 599 & Patents, drug labels & UMLS \\
MedMentions (MedM) \cite{mohan2019medmentionslargebiomedicalcorpus}  & 8,863 & PubMed abstracts & UMLS \\
COMETA (COM) \cite{basaldella2020cometa}  & 4,226 & Reddit, Twitter posts & SNOMED CT \\
MIMIC-IV-EL (MIM) \cite{davidson2025snomed}  & 20,404 & ICU discharge notes & SNOMED CT \\
TAC-ADR (ADR) \cite{roberts2017overview} & 8,157 & FDA drug labels & MedDRA, RxNorm \\
CADEC (CAD) \cite{karimi2015cadec}  & 7,453 & Patient forum posts & SNOMED CT, MedDRA \\
NCBI Disease (NCBI) \cite{dougan2014ncbi}  & 959 & PubMed abstracts & MeSH, OMIM \\
BC5CDR (CDR) \cite{li2016biocreative}  & 4,883 & PubMed abstracts & MeSH \\
\midrule
BioC-GS (BioC) \cite{weissenbacher2023phenoid}  & 480 & Genetics clinical notes & HPO \\
GSC+ (GSC) \cite{gsc+}  & 1,440 & Biomedical abstracts & HPO \\
\bottomrule
\end{tabular}
}
\caption{Evaluation datasets. }
\label{tab:datasets}
\end{table*}

\subsection{Baselines}
We first compare our method against other LLM-based methods, REAL \cite{shlyk2024real} and AutoPCR \cite{tao2025autopcr}. AutoPCR is straightforward to adapt to cross-ontology evaluation; we simply build the ontology embedding for each target ontology using the default SapBERT model \cite{liu2021self}, and the other parts of the pipeline are ontology-agnostic (Stanza \cite{zhang2021biomedical} BioNER for Mention Detection and LLM candidate selection). The REAL method can also be adapted by first building the ontology embedding using the method's default embeddings model (all-MiniLM-L6-v2 \cite{wang2020minilm}) and creating a target entity description that fits each dataset's target entities for the mention detection step.

We also compare our method against rule-based approaches: NCBO \cite{Jonquet2009OBA}, QuickUMLS \cite{soldaini2016quickumls}, and FastHPOCR \cite{groza2024fasthpocr}. We chose these methods because they obtain high performance on some of the included datasets, and were each designed with a different target ontology. Adapting these methods to the cross-ontology setting is also straightforward; the ontology file simply needs to be swapped, but a drop in performance in domains outside the initial scope of the methods is to be expected. 

Supervised methods are also used as a baseline; however, these cannot be directly adapted to the cross-ontology setting. We select the best performing end-to-end methods with available code for reproduction across the different test datasets. We select PhenoTagger \cite{luo2021phenotagger} as a strong performer on the phenotype concept recognition tasks, TaggerOne \cite{leaman2016taggerone} which was proposed on the BC5CDR and NCBI Disease datasets, DNorm \cite{leaman2013dnorm}, also proposed on NCBI Disease, KIRIs, the winning team and corresponding system from the SNOMEDCT MIMIC-IV EL Challenge \cite{davidson2025snomed}, and CONORM \cite{yazdani2023conorm}, top-performing system from the SMM4H 2023 ADE shared task and tested on CADEC. Implementation details for these baselines are presented in Appendix \ref{app:baselines}. These represent baseline performance for the in-distribution training scenario. We note that many of these datasets have not been tested on the end-to-end Concept Recognition task and, as such, do not have a supervised approach as a baseline.


\subsection{Backbone Models}
We run all LLM-based approaches using the Qwen3.8-27B model \cite{qwen38} as backbone. We also run a comparison of these methods using the Qwen3-8B model \cite{yang2025qwen3} in Section \ref{sec:analysis}. All backbone models are run using default temperature (0.7). Exact model checkpoints, sampling parameters, inference stack, computational setup and hyperparameters are detailed in Appendix \ref{app:impl}.

\begin{table*}[t]
\centering
\normalsize
\setlength{\tabcolsep}{2.0pt}
\renewcommand{\arraystretch}{1.08}

\begin{tabular*}{\textwidth}{@{\extracolsep{\fill}}
    l
    *{11}{c}
    @{\hspace{4pt}}|@{\hspace{4pt}}
    c@{\hspace{1pt}}
    c
    @{}}

\toprule

& \multicolumn{9}{c}{MedPath}
& \multicolumn{2}{c}{Phenotype}
& \multicolumn{2}{c}{Overall} \\

\cmidrule(lr){2-10}
\cmidrule(lr){11-12}
\cmidrule(lr){13-14}

Model
& SC & MAN & MedM & COM
& ADR & CAD & MIM & NCBI & CDR
& BioC & GSC
& Avg & Rank \\

\midrule


\makecell[l]{\hspace{-4pt}\textit{LLM-based}}
& & & & & & & & & & & & & \\

\hspace{15pt}UMCR
& \textbf{0.56}
& \textbf{0.61}
& \textbf{0.44}
& \textbf{0.34}
& \textbf{0.66}
& \underline{0.27}
& \underline{0.35}
& 0.50
& \textbf{0.58}
& \textbf{0.70}
& 0.60
& \textbf{0.51}
& \textbf{1.18}
\\

\hspace{15pt}AutoPCR
& 0.40
& \underline{0.46}
& \underline{0.35}
& 0.15
& 0.38
& 0.23
& 0.19
& 0.40
& 0.43
& \underline{0.69}
& \underline{0.69}
& \underline{0.40}
& \underline{2.55}
\\

\hspace{15pt}REAL
& 0.26
& 0.41
& 0.11
& \underline{0.30}
& 0.10
& 0.26
& 0.19
& 0.27
& 0.26
& 0.64
& 0.36
& 0.29
& 4.00
\\

\midrule


\makecell[l]{\hspace{-4pt}\textit{Rule-based}}
& & & & & & & & & & & & & \\

\hspace{15pt}NCBO
& 0.45
& 0.25
& 0.00
& 0.11
& \underline{0.52}
& 0.22
& 0.15
& 0.18
& 0.26
& 0.46
& 0.56
& 0.29
& 4.27
\\

\hspace{15pt}QuickUMLS
& 0.45
& 0.21
& 0.07
& 0.11
& 0.29
& 0.21
& 0.15
& 0.38
& 0.29
& 0.43
& 0.42
& 0.27
& 4.55
\\

\hspace{15pt}FastHPOCR
& \underline{0.46}
& 0.23
& 0.18
& 0.07
& 0.35
& 0.08
& 0.11
& 0.32
& 0.35
& 0.57
& \textbf{0.73}
& 0.31
& 4.00
\\

\midrule


\makecell[l]{\hspace{-4pt}\textit{Supervised}}
& & & & & & & & & & & & & \\

\hspace{15pt}PhenoTagger
& $-$
& $-$
& $-$
& $-$
& $-$
& $-$
& $-$
& $-$
& $-$
& \cellcolor{black!10}0.65
& \cellcolor{black!10}0.60
& $-$
& $-$
\\

\hspace{15pt}TaggerOne
& 0.30
& $-$
& $-$
& $-$
& $-$
& $-$
& $-$
& \cellcolor{black!10}\textbf{0.68}
& \cellcolor{black!10}\underline{0.55}
& $-$
& $-$
& $-$
& $-$
\\

\hspace{15pt}DNorm
& 0.18
& $-$
& $-$
& $-$
& $-$
& $-$
& $-$
& \cellcolor{black!10}\underline{0.60}
& $-$
& $-$
& $-$
& $-$
& $-$
\\

\hspace{15pt}KIRIs
& $-$
& $-$
& $-$
& $-$
& $-$
& $-$
& \cellcolor{black!10}\textbf{0.55}
& $-$
& $-$
& $-$
& $-$
& $-$
& $-$
\\

\hspace{15pt}CONORM
& $-$
& $-$
& $-$
& $-$
& 0.15
& \cellcolor{black!10}\textbf{0.65}
& $-$
& $-$
& $-$
& $-$
& $-$
& $-$
& $-$
\\

\bottomrule
\end{tabular*}

\caption{Overlap Linking F1 Score. Bold marks the best measured score per dataset, underline the second best. Grey cells mark in-distribution results. \textit{Avg} represents the average score across the datasets ($\uparrow$), and \textit{Rank} is the average ranking per dataset for each model ($\downarrow$). Qwen3.8-27B is the backbone model for all LLM-based methods.}
\label{tab:main}
\end{table*}

\subsection{Ablations}

To measure the impact of the components of our proposed pipeline, we conduct ablation experiments on four selected datasets (ShAReCLEF, NCBI Disease, BioC-GS, and GSC+), using their respective development sets\footnote{The training set is used for ShAReCLEF, see \ref{sec:datasets}.}. This subset contains 11,068 sentences and a variety of input text type and ontologies. It also allows direct comparison to the other LLM-based methods in their target domain. All ablations were run after the main evaluation experiments and did not influence our pipeline design, but rather served to understand the impact of each component.

We first measure the impact of the guideline compilation step on the pipeline, by comparing three guideline versions; \texttt{compiled guidelines}, which we manually encode in the JSON format (see Section \ref{sec:method}), \texttt{bare guidelines}, where we use a single phrase to describe the target entities, and static instructions for linking and filtering, and \texttt{full guidelines}, where we give the model full dataset guidelines under their initial formats (PDFs, html, etc.) Examples of each guideline style are presented in Appendix \ref{app:guideline_compilation}.

Next, we ablate the effect of the different stages of the pipeline by measuring mention detection and entity linking performance at each stage. We note that Mention Detection (Stage 2) does not have entity linking because its output consists only of spans, and Entity Linking in Stage 3 can impact mention detection through the \texttt{adjust\_span} tool.

We then measure the specific impact of the agentic linking step by comparing it against a fixed linking module, where we first retrieve candidates for the given mention (\texttt{search} tool), return their information (\texttt{get\_concept} tool), and prompt the model to select the best candidate. For this experiment, we do not run the filtering stage to isolate the impact of the linking phase. We also investigate how this module makes use of the available tools.

\section{Results}
\label{sec:results}
\subsection{Medical Concept Recognition}

Evaluation results on MedPath and the phenotype datasets are presented in Table \ref{tab:main}. We report overlap-matching F1 Score for end-to-end Concept Recognition. Complete results for mention detection and exact-matching metrics are provided in Appendix \ref{app:res} and lead to the same overall conclusions. UMCR achieves the strongest results overall, obtaining the highest score in 7 of the 11 configurations, with an average F1 Score of 0.51 and an average rank of 1.18, showing it can adapt across the datasets. 

UMCR compares favorably to supervised approaches, despite not relying on dataset-specific supervised training. In particular, it outperforms the supervised systems reported for BioC-GS (0.70 F1 Score) and BC5CDR (0.58 F1 Score), while remaining competitive with the strongest supervised results on the other datasets. The main exception is CADEC, where CONORM achieves a substantially higher score of 0.65 compared with 0.27 for UMCR, in second place. The lowest UMCR performance is seen on datasets with thin guidelines (MIM), particular task setups (COM) or high guideline specificity (CAD). When applying supervised approaches to out-of-distribution datasets using the same target ontologies, we see severe drops in performance, like CONORM's 0.15 F1 score on the TAC-ADR dataset. This illustrates the lack of generalizability from these trained systems.

Comparing with the other LLM-based approaches further highlights the robustness of UMCR. UMCR consistently outperforms REAL across all evaluated datasets and achieves higher scores than AutoPCR on 9 of 11 datasets. On the two phenotype datasets for which REAL and AutoPCR were specifically developed, UMCR outperforms REAL on both datasets and achieves a higher score than AutoPCR on BioC-GS (0.70 vs. 0.69), while remaining competitive on GSC+ (0.60 vs. 0.69).

UMCR also substantially improves over the rule-based baselines overall, with an average difference of +19 F1 Score points across all datasets over the three baselines. It achieves a higher F1 Score than all rule-based methods on 10 of the 11 datasets, with GSC+ being the only exception, where FastHPOCR obtains 0.73 compared with 0.60 for UMCR. The particularly large differences on several MedPath datasets illustrate the limitations of rule-based methods when applied across ontologies and input text types.

These results demonstrate that UMCR is a competitive and consistent approach across ontologies, domains and text types. UMCR allows for broad generalization while remaining competitive with supervised approaches which necessitate ontology and dataset-specific training data, a requirement that is often impractical.

\subsection{Ablation Results}
\label{sec:ablations_res}

\paragraph{Guideline Compilation} Table \ref{tab:guideline} shows results of the guideline compilation ablation experiment, where we test three different guideline compilation strategies. We observe that our \texttt{compiled} guidelines obtain performance gains over the other two versions on all tested datasets, with average gains of 4 F1 Score points against the \texttt{bare} version, and 3.25 F1 Score points against the \texttt{full} version. We hypothesize that the bare guidelines do not contain enough detailed information on the task requirements, while the full guidelines might be more difficult for the model to use compared to our compiled JSON format.

\begin{table}[h]
\centering

\begin{tabular}{@{}lrrr@{}}
\toprule
Dataset & \texttt{bare} & \texttt{compiled} & \texttt{full} \\
\midrule
ShAReCLEF & 0.48 & \textbf{0.54} & 0.50 \\
NCBI Disease & 0.35 & \textbf{0.39} & 0.38 \\
BioC-GS & 0.62 & \textbf{0.64} & 0.62 \\
GSC+ & 0.54 & \textbf{0.59} & 0.53 \\
\bottomrule
\end{tabular}
\caption{Guideline Compilation Ablation: overlap linking F1 Score after Stage 3. Bold marks the best configuration.}
\label{tab:guideline}
\end{table}

\paragraph{Stage Ablation} We also study how the different stages impact performance on both the Mention Detection task and the Entity Linking task, with results presented in Table \ref{tab:stages}. For Mention Detection, the results show small improvements from Stage 2 to Stage 3 from the \texttt{adjust\_span} tool, and larger gains from the filtering done by the last stage, although one dataset, ShAReCLEF, does not see improvements beyond the initial Stage 2 performance. Entity Linking sees gains on all datasets from Stage 3's agentic linking to the filtered, final output, with an average improvement of 5.25 F1 Score points.

\begin{table}[h]
\centering

\begin{tabular}{@{}lccc@{}}
\toprule
Dataset & Stage 2 & Stage 3 & Stage 4 \\
\midrule
\multicolumn{4}{@{}l}{\textbf{Overlap Mention Detection F1 Score}} \\
ShAReCLEF    & 0.70 & 0.70 & 0.70 \\
NCBI Disease & 0.63 & 0.67 & 0.79 \\
BioC-GS      & 0.81 & 0.82 & 0.85 \\
GSC+         & 0.63 & 0.63 & 0.65 \\
\midrule
\multicolumn{4}{@{}l}{\textbf{Overlap Linking  F1 Score}} \\
ShAReCLEF    & $-$   & 0.54 & 0.56 \\
NCBI Disease & $-$   & 0.39 & 0.52 \\
BioC-GS      & $-$   & 0.64 & 0.71 \\
GSC+         & $-$   & 0.58 & 0.59 \\
\bottomrule
\end{tabular}
\caption{Stage Ablation Results. Stage 2: Mention Detection, Stage 3: Agentic Linking, Stage 4: Filtering.}
\label{tab:stages}
\end{table}

\paragraph{Agentic Linking} Next, Table \ref{tab:linking} shows the results of the agentic linking ablation, comparing our Stage 3 module to a version with a fixed path. The fixed path first retrieves candidates from the given mention, then tasks the model with selecting the best candidate, with the compiled guidelines as context. Performance is degraded on all datasets by limiting the actions and not allowing the model to choose its ontology exploration strategy, with an average performance loss of 7.25 F1 Score points.

\begin{table}[h]
\centering

\begin{tabular}{@{}lrrr@{}}
\toprule
Dataset & Agentic Linking & Fixed \\
\midrule
ShAReCLEF & 0.54 & 0.49 \\
NCBI Disease & 0.39 & 0.31 \\
BioC-GS & 0.64 & 0.52 \\
GSC+ & 0.59 & 0.55\\
\bottomrule
\end{tabular}
\caption{Linking Type Ablation: Overlap linking F1 Score after Stage 3.}
\label{tab:linking}
\end{table}
We further investigate Stage 3's agentic linking, studying how useful exploration of the ontology beyond the \texttt{search} and \texttt{get\_concept} tools is. Our hypothesis is that \textit{graph exploration}, meaning the movement-based tools (\texttt{get\_parents}, \texttt{get\_children}, \texttt{get\_siblings}), should help the model identify the correct concept. To study this, we devised three Stage 3 prompt variants. The initial one, named 'Optional' mentions that graph exploration is optional for the linking stage. The second version, 'Encouraged', mentions graph exploration as a beneficial step. The third (and default) prompt, 'Mandatory', explicitly forces the model through instructions to use these different tools before committing to a concept. Prompts are shown in Appendix \ref{app:prompts}.

\begin{figure}[h]
\centering
\includegraphics[width=\columnwidth]{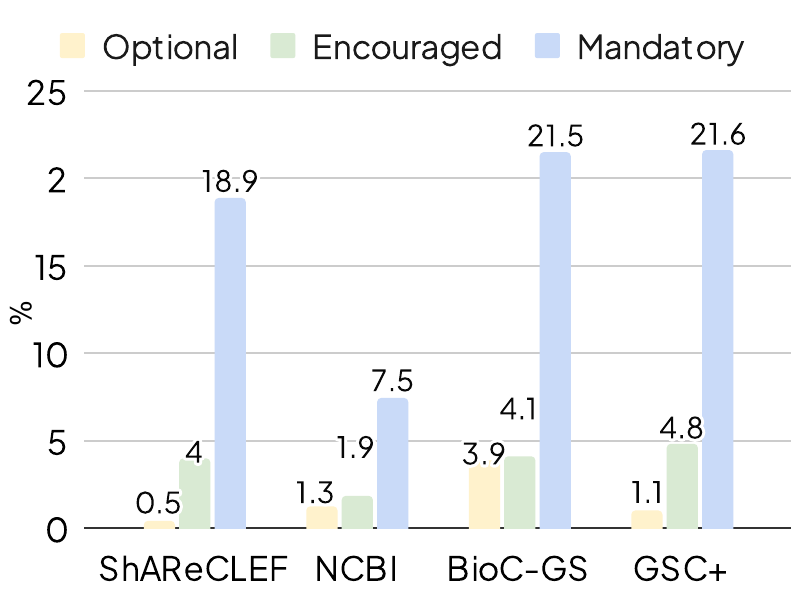}
\caption{Percentage of graph exploration tool calls per mention.}
\label{fig:graph_hops}
\end{figure}

\begin{table}[h]
\centering

\setlength{\tabcolsep}{2pt}
\resizebox{\columnwidth}{!}{%
\begin{tabular}{@{}lrrr@{}}
\toprule
Dataset & \textit{Optional} & \textit{Encouraged} & \textit{Mandatory} \\
\midrule
ShAReCLEF & 0.50 & 0.51 & 0.54 \\
NCBI Disease & 0.38 & 0.39 & 0.39 \\
BioC-GS & 0.59 & 0.60 & 0.64 \\
GSC+ & 0.56 & 0.58 & 0.59 \\
\bottomrule
\end{tabular}
}
\caption{Agentic Linking Ablation (Prompt Choice): Overlap linking F1 Score after Stage 3}
\label{tab:agentic_linking}
\end{table}



Figure \ref{fig:graph_hops} shows the percentage of Stage 3 calls that are graph exploration steps. We see that the second prompt variation slightly increases this percentage, but that forcing the model to explore before committing has a far greater impact. Looking at performance resulting from these changes, in Table \ref{tab:agentic_linking}, we observe a similar phenomenon, with slight increase in performance from the 'Encouraged' prompt, and greater gains from the forced exploration.

Overall, these ablations experiments show the soundness of our pipeline design, with each stage improving model performance. In particular, we show that using the guidelines in our compiled format and allowing agentic exploration of the ontology for the linking stage both lead to large performance gains.

\subsection{Backbone Model}
\label{sec:analysis}
\begin{figure}[h]
\centering
\includegraphics[width=\columnwidth]{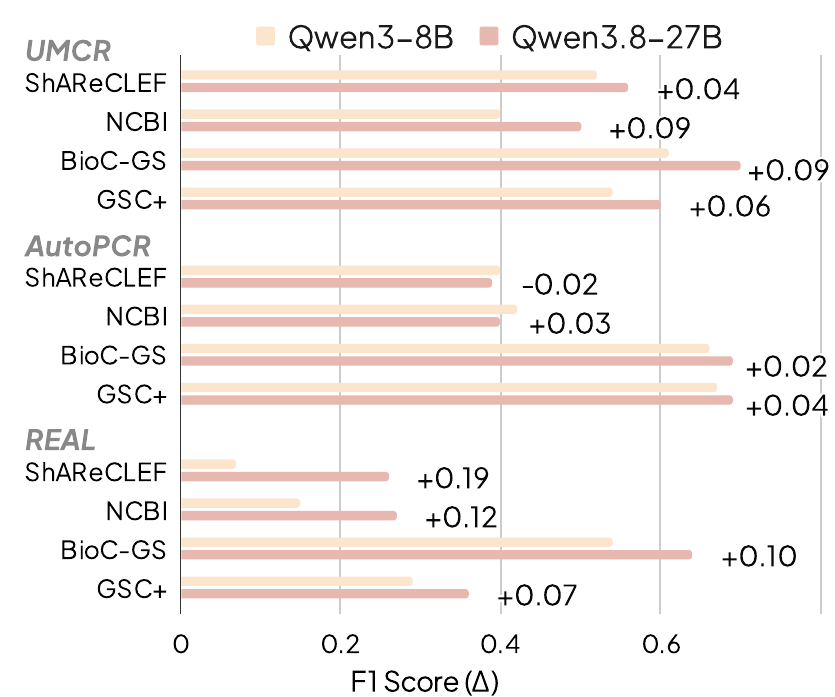}
\caption{Backbone LLM Ablation: Overlap linking F1 Score. $\Delta$ measures difference between both models.}
\label{fig:backend}
\end{figure}

Finally, we evaluate the role of the backbone model in the different LLM-based approaches, with results presented in Figure \ref{fig:backend}. We observe that for UMCR, using a stronger and larger model yields improvements across all datasets. We also note that AutoPCR is less impacted by this improved model, even degrading in performance on some datasets. The REAL approach sees the largest increase in performance; however, its performance with the 27B model is still below that of our model in both configurations. This highlights our method's effective use of the agentic capabilities of the backbone model without relying on the parametric memory of the LLM.

\section{Conclusion}
In this study, we present the UMCR model, a pipeline for medical concept recognition that allows extraction and linking towards any medical ontology, without requiring training or labeled data. We show that UMCR is superior to other LLM and rule-based approaches and competes with supervised systems across datasets. We ablate the different choices in the pipeline, concluding that guideline compilation and agentic linking both help in reaching this performance level. Overall, UMCR represents a strong approach for general medical concept recognition, with potential to bring  interoperability between healthcare services.

\section*{Limitations}

We first highlight the inherent challenges of information extraction as a general task, particularly in designing methods that generalize across ontologies and text types. While UMCR achieves the best average performance, it remains low at 0.51 overlap linking F1 Score. This highlights that further progress can be made to improve performance, particularly towards datasets that have not often been evaluated on this end-to-end task.

One of the main limitations of our approach is the cost of running the pipeline, compared to rule-based and other LLM-based approaches. Because of the agentic linking module, our method requires many LLM calls, making it slower and more compute-intensive. This is detailed further in Appendix \ref{app:impl}. 

Another point to consider is the hierarchical paths provided with the MedPath dataset: these metrics allow for more fine-grained evaluation of how close an output concept is to the gold label, within a given ontology. Including these metrics could allow for further improvements of our method and is left for future work.

We also designed our pipeline with interpretability in mind: the mention detection module produces a \textit{confidence score} along with predicted spans, and the entity linking and filtering parts of the model produce \textit{reasons} for each decision. These are not used in the current setup but could be introduced to better control the approach. 


Finally, one of the included datasets, Mantra-GSC, provides multilingual annotations and input text, but we only test our method on the English language.

\bibliography{ref}

\appendix
\section*{Appendix}

\section{Implementation details}
\label{app:impl}

\paragraph{Compute.}
The pipeline is developed with vLLM to serve inference. All experiments were run on NVIDIA H100 GPUs, apart from the rule-based and supervised methods which were run on CPU (Intel\textsuperscript{®} Xeon\textsuperscript{®} Gold 6448Y Processor). 

\paragraph{Processing time.}

We mention the higher processing time of the method as one of its limitations. We measure this on the four ablation datasets by computing the average number of calls per sentence, and we measure an average per-call latency of 5 seconds. This gives us estimates of per-sentence processing time for each of those datasets, depending on mention density. These estimates are presented in Table \ref{tab:dataset_timing}. This does not take into account the startup time, which is largely attributed to initial model loading.

\begin{table}[h]
\centering
\begin{tabular}{lcc}
\hline
\textbf{Dataset} & \textbf{Calls/sent.} & 
\textbf{$\approx$ s/sent} \\
\hline
ShAReCLEF   & 0.4 & $\sim$2s  \\
NCBI Disease & 3.1 & $\sim$16s \\
BioC-GS     & 3.6 & $\sim$18s \\
GSC+        & 2.6 & $\sim$13s \\

\hline
\end{tabular}
\caption{Calls and estimated processing time per sentence.}
\label{tab:dataset_timing}
\end{table}

\paragraph{Hyperparameters.}
All LLM-based models used the Qwen3.8-27B model for inference. Hyperparameters are presented in Table \ref{tab:qwen38}. \texttt{max\_model\_len} is reduced to fit KV cache on H100. Other parameters for UMCR include: 8 max steps for agentic linking, and 10 candidates for the \texttt{search} operation. 

\begin{table}[h]
\centering
\begin{tabular}{@{}ll@{}}
\toprule
Parameter & Value \\
\midrule
Precision & BF16 \\
Thinking mode & off \\
\texttt{temperature} & 0.7 \\
\texttt{top\_p} & 0.80 \\
\texttt{top\_k} & 20 \\
\texttt{min\_p} & 0.0 \\
\texttt{presence\_penalty} & 1.5 \\
\texttt{repetition\_penalty} & 1.0 \\
\texttt{max\_tokens} (S2/S3/S4) & 2048/1024/2048 \\
\texttt{gpu-memory-utilization} & 0.85 \\
\texttt{max-model-len} & 32k \\
\bottomrule
\end{tabular}
\caption{Qwen3.8-27B generation configuration.}
\label{tab:qwen38}
\end{table}

\paragraph{Embeddings and Retrieval.}
For retrieval, every ontology concept (names and synonyms) is embedded once with Qwen3-Embedding-0.6B into L2-normalized vectors, supporting cosine search via inner product. At query time, the candidate string is embedded with the same model, and top 50 candidates are retrieved using dense vector search. We also retrieve exact matches between the candidate string and the ontology concept names and synonyms, as well as the top 50 BM25 matches. These three candidate pools are then merged with reciprocal rank fusion (RRF, k=60), returning the top 10 concepts. Results are cached per query so they can be reused.  Embeddings are stored in raw NumPy \texttt{memmap}.

\paragraph{Datasets.} Some datasets have annotations from more than one target ontology (TAC-ADR, CADEC, NCBI Disease). For those, we run each model once on each ontology, and merge the results by first using the output from the main ontology, and adding results from the others if a given mention is not covered by the main run. Main ontologies in that case are determined manually using the annotation guidelines and the number of annotations per ontology on the respective dataset's training split.

\section{Agentic Linking Tools}
\label{app:stage3-tools}

Table \ref{tab:stage3-tools} details the functions and signatures of the available tools for Stage 3's linking process.

\begin{table*}[h]
\centering

\resizebox{\textwidth}{!}{
\begin{tabular}{@{}lll@{}}
\toprule
Tool & Arguments & Returns \\
\midrule
\texttt{search} & \texttt{query}, \texttt{k} (=10) & Top-k candidate concepts (exact matching + vector search + BM25) \\
\texttt{get\_concept} & \texttt{id} & Name, definition, synonyms, and semantic types of \texttt{id} concept \\
\texttt{get\_parents} & \texttt{id} & Parent concepts of a candidate \\
\texttt{get\_children} & \texttt{id} & Children concepts of a candidate \\
\texttt{get\_siblings} & \texttt{id} & Siblings concepts of a candidate \\
\texttt{adjust\_span} & \texttt{text} & Refines span boundaries of current mention \\
\texttt{commit} & \texttt{id}, \texttt{confidence} & Accept a candidate \texttt{id} for the mention and end the loop \\
\texttt{unlink} & \texttt{reason} & Reject all candidates (no eligible concept fits) and end the loop \\
\bottomrule
\end{tabular}
}
\caption{Tools available to the Stage 3 Agentic Linking Module.}
\label{tab:stage3-tools}
\end{table*}

\section{Baselines}
\label{app:baselines}
This section details each baseline method and how it was adapted to the cross-ontology evaluation.

\paragraph{AutoPCR.} Three stages: rule-based and neural mention extraction (BioNER with syntax-based decomposition), SapBERT candidate retrieval over an embedded ontology index, and LLM linking over candidate identifiers, names, definitions, and synonyms \cite{tao2025autopcr}. We build one SapBERT index per target ontology, and run extraction with Stanza neural++ settings, linking uncertain cases with Qwen3.8-27B.

\paragraph{REAL.} Mention extraction with an LLM, dense retrieval of top-$k$ ontology candidates (ChromaDB with all-MiniLM-L6-v2 embeddings), and LLM grounding that keeps only high-confidence links \cite{shlyk2024real}. We build one retrieval index per target ontology, write dataset-specific mention-detection prompts naming the target entities, and run grounding with Qwen3.8-27B ($k=3$).

\paragraph{NCBO Annotator.} We use the BioPortal API \cite{Jonquet2009OBA}. We query the API with semantic-type filters devised from the annotation guidelines (e.g., subset of 11 disorder semantic types for ShAReCLEF).

\paragraph{QuickUMLS.} String matching against a SimString index built from UMLS 2026AA \cite{soldaini2016quickumls}. We build per-dataset index subsets reflecting the target ontology for each dataset. We use the default matching threshold of 0.7.

\paragraph{FastHPOCR.} We use the default \texttt{expanded} variant and index each target ontology using morpheme-cluster dictionaries built per ontology \cite{groza2024fasthpocr}.

\paragraph{PhenoTagger.} Hybrid dictionary and BioBERT/PubMedBERT tagging trained by distant supervision on the target ontology, requiring no manual labels \cite{luo2021phenotagger}. We use the pretrained PubMedBERT model.

\paragraph{TaggerOne.} Joint neural NER and normalization model \cite{leaman2016taggerone}. We run its shipped NCBI Disease and BC5CDR models without retraining. We apply the method out of domain to ShAReCLEF with the NCBI model and convert the identifiers to the SNOMED ontology using the UMLS links.

\paragraph{DNorm.} BANNER CRF mention detection with pairwise learning-to-rank normalization over the MEDIC ontology \cite{leaman2013dnorm}. We run its shipped NCBI models without retraining and map MEDIC identifiers to MeSH and OMIM identifiers using UMLS mapping. We apply the method out-of-distribution to the ShAReCLEF dataset.

\paragraph{KIRIs.} Dictionary-based winner of the SNOMED CT entity-linking challenge \cite{davidson2025snomed}. We use all default hyperparameters.

\paragraph{CONORM.} We retrain the NER (MPNet) plus bi-encoder normalization to MedDRA ontology concepts \cite{yazdani2023conorm}. Trained on the CADEC training set and run on its test split. We reuse this model out-of-distribution on the TAC-ADR dataset, as it already outputs MedDRA identifiers.

\section{Tool calling Analysis}
\label{app:tool_analysis}
Table~\ref{tab:tool-use} summarizes Stage 3 tool use on the four ablation datasets for the default configuration (Qwen3.8-27B). A significant proportion of mentions do not enter the agent loop: exact synonym matches skip Stage 3 (64--87\% of mentions), and this step handles the remaining mentions, at an average of 3.6 to 6.0 calls per mention within the 8-step budget. Calls concentrate on \texttt{search}, \texttt{get\_concept}, and terminal \texttt{commit}; graph navigation (\texttt{get\_parents}, \texttt{get\_children}, \texttt{get\_siblings}) accounts for 7 to 22\% of calls under the default exploration-forcing prompt. \texttt{adjust\_span} is rarely used by the model, leaving the candidate spans untouched. Further analysis shows that correctly linked mentions resolve quickly (under one step on average) while incorrectly linked mentions consume more steps before committing or unlinking.
\begin{table*}[h]
\centering
\caption{Stage 3 Tool Use per dataset. Agent \% = percentage of Stage 2 mentions that enter Stage 3; Calls = Average \# of calls per mention; Last four columns are tool call proportion over Stage 3.}
\label{tab:tool-use}
\begin{tabular}{@{}lrrrrrr@{}}
\toprule
Dataset & Agent \% & Calls & \texttt{search} & \texttt{get\_concept} & Graph & \texttt{commit/unlink} \\
\midrule
ShAReCLEF & 20.2 & 5.26 & 27.6 & 37.9 & 18.9 & 15.6 \\
NCBI Disease & 13.0 & 3.63 & 34.7 & 30.3 & 7.5 & 27.6 \\
BioC-GS & 35.9 & 4.81 & 34.6 & 26.3 & 21.5 & 17.6 \\
GSC+ & 24.7 & 6.01 & 47.3 & 20.4 & 21.6 & 10.8 \\
\bottomrule
\end{tabular}
\end{table*}

\section{Full results}
\label{app:res}

This section presents the full results for both levels of evaluation, with both evaluation metrics; Table \ref{tab:ner} shows \textit{Overlap} \textbf{Mention Detection} F1 Score, and Table \ref{tab:ner-exact} shows \textit{Exact Matching} \textbf{Mention Detection} F1 Score, while Table \ref{tab:main-exact} shows Exact Matching \textbf{Linking} F1 Score.

These results lead to the same overall conclusions as our analysis in Section \ref{sec:results}, where we highlight the strong performance of our UMCR method across target vocabularies and source text. UMCR outperforms the other tested LLM and rule-based methods, and competes with supervised methods without requiring any annotated text data.

\begin{table*}[t]
\centering
\normalsize
\setlength{\tabcolsep}{2.0pt}
\renewcommand{\arraystretch}{1.08}

\begin{tabular*}{\textwidth}{@{\extracolsep{\fill}}
    l
    *{11}{c}
    @{\hspace{4pt}}|@{\hspace{4pt}}
    c@{\hspace{1pt}}
    c
    @{}}

\toprule

& \multicolumn{9}{c}{MedPath}
& \multicolumn{2}{c}{Phenotype}
& \multicolumn{2}{c}{Overall} \\

\cmidrule(lr){2-10}
\cmidrule(lr){11-12}
\cmidrule(lr){13-14}

Model
& SC & MAN & MedM & COM
& ADR & CAD & MIM & NCBI
& CDR & BioC & GSC
& Avg & Rank \\

\midrule


\makecell[l]{\hspace{-4pt}\textit{LLM-based}}
& & & & & & & & & & & & & \\

\hspace{15pt}UMCR
& \textbf{0.73}
& \textbf{0.73}
& \textbf{0.66}
& \textbf{0.49}
& \textbf{0.74}
& 0.48
& \underline{0.61}
& 0.78
& \underline{0.82}
& \textbf{0.85}
& \underline{0.67}
& \textbf{0.69}
& \textbf{1.18}
\\

\hspace{15pt}AutoPCR
& 0.52
& \underline{0.64}
& \underline{0.64}
& 0.22
& 0.48
& 0.37
& 0.38
& 0.70
& 0.65
& 0.78
& \textbf{0.74}
& \underline{0.56}
& \underline{3.09}
\\

\hspace{15pt}REAL
& 0.39
& 0.60
& 0.20
& \underline{0.46}
& 0.17
& \underline{0.61}
& 0.34
& 0.42
& 0.40
& 0.75
& 0.44
& 0.43
& 4.45
\\

\midrule


\makecell[l]{\hspace{-4pt}\textit{Rule-based}}
& & & & & & & & & & & & & \\

\hspace{15pt}NCBO
& 0.59
& 0.47
& 0.03
& 0.20
& \underline{0.63}
& 0.39
& 0.34
& 0.51
& 0.70
& 0.57
& 0.60
& 0.46
& 4.00
\\

\hspace{15pt}QuickUMLS
& \underline{0.63}
& 0.46
& 0.40
& 0.22
& 0.44
& 0.42
& 0.40
& 0.63
& 0.53
& 0.51
& 0.45
& 0.46
& 3.91
\\

\hspace{15pt}FastHPOCR
& 0.60
& 0.44
& 0.51
& 0.12
& 0.48
& 0.22
& 0.29
& 0.45
& 0.42
& 0.68
& \textbf{0.74}
& 0.45
& 4.45
\\

\midrule


\makecell[l]{\hspace{-4pt}\textit{Supervised}}
& & & & & & & & & & & & & \\

\hspace{15pt}PhenoTagger
& $-$
& $-$
& $-$
& $-$
& $-$
& $-$
& $-$
& $-$
& $-$
& \cellcolor{black!10}\underline{0.79}
& \cellcolor{black!10}0.65
& $-$
& $-$
\\

\hspace{15pt}TaggerOne
& 0.62
& $-$
& $-$
& $-$
& $-$
& $-$
& $-$
& \cellcolor{black!10}\textbf{0.92}
& \cellcolor{black!10}\textbf{0.89}
& $-$
& $-$
& $-$
& $-$
\\

\hspace{15pt}DNorm
& 0.34
& $-$
& $-$
& $-$
& $-$
& $-$
& $-$
& \cellcolor{black!10}\underline{0.84}
& $-$
& $-$
& $-$
& $-$
& $-$
\\

\hspace{15pt}KIRIs
& $-$
& $-$
& $-$
& $-$
& $-$
& $-$
& \cellcolor{black!10}\textbf{0.68}
& $-$
& $-$
& $-$
& $-$
& $-$
& $-$
\\

\hspace{15pt}CONORM
& $-$
& $-$
& $-$
& $-$
& 0.18
& \cellcolor{black!10}\textbf{0.62}
& $-$
& $-$
& $-$
& $-$
& $-$
& $-$
& $-$
\\


\bottomrule
\end{tabular*}

\caption{Overlap Mention-Detection F1 Score. Bold marks the best measured score per dataset, underline the second best. Grey cells mark in-distribution results. \textit{Avg} represents the average score across the datasets, and \textit{Rank} is the average ranking per dataset for each model.}
\label{tab:ner}
\end{table*}

\begin{table*}[t]
\centering
\normalsize
\setlength{\tabcolsep}{2.0pt}
\renewcommand{\arraystretch}{1.08}

\begin{tabular*}{\textwidth}{@{\extracolsep{\fill}}
    l
    *{11}{c}
    @{\hspace{4pt}}|@{\hspace{4pt}}
    c@{\hspace{1pt}}
    c
    @{}}

\toprule

& \multicolumn{9}{c}{MedPath}
& \multicolumn{2}{c}{Phenotype}
& \multicolumn{2}{c}{Overall} \\

\cmidrule(lr){2-10}
\cmidrule(lr){11-12}
\cmidrule(lr){13-14}

Model
& SC & MAN & MedM & COM
& ADR & CAD & MIM & NCBI
& CDR & BioC & GSC
& Avg & Rank \\

\midrule


\makecell[l]{\hspace{-4pt}\textit{LLM-based}}
& & & & & & & & & & & & & \\

\hspace{15pt}UMCR
& \textbf{0.61}
& \textbf{0.66}
& \underline{0.51}
& \textbf{0.42}
& \textbf{0.67}
& 0.28
& \underline{0.42}
& 0.58
& \underline{0.75}
& \textbf{0.62}
& 0.62
& \textbf{0.56}
& \textbf{1.45}
\\

\hspace{15pt}AutoPCR
& 0.45
& 0.54
& \textbf{0.53}
& 0.18
& 0.41
& 0.24
& 0.31
& 0.51
& 0.59
& \textbf{0.62}
& \underline{0.70}
& \underline{0.46}
& \underline{2.55}
\\

\hspace{15pt}REAL
& 0.32
& 0.48
& 0.15
& \underline{0.40}
& 0.15
& \underline{0.47}
& 0.25
& 0.31
& 0.32
& 0.47
& 0.38
& 0.34
& 4.36
\\

\midrule


\makecell[l]{\hspace{-4pt}\textit{Rule-based}}
& & & & & & & & & & & & & \\

\hspace{15pt}NCBO
& 0.50
& 0.41
& 0.03
& 0.13
& \underline{0.53}
& 0.21
& 0.25
& 0.36
& 0.66
& 0.41
& 0.56
& 0.37
& 4.09
\\

\hspace{15pt}QuickUMLS
& \underline{0.53}
& 0.34
& 0.27
& 0.16
& 0.32
& 0.22
& 0.28
& 0.48
& 0.45
& 0.40
& 0.43
& 0.36
& 4.09
\\

\hspace{15pt}FastHPOCR
& 0.49
& 0.34
& 0.42
& 0.09
& 0.41
& 0.12
& 0.21
& 0.31
& 0.38
& 0.50
& \textbf{0.73}
& 0.36
& 4.45
\\

\midrule


\makecell[l]{\hspace{-4pt}\textit{Supervised}}
& & & & & & & & & & & & & \\

\hspace{15pt}PhenoTagger
& $-$
& $-$
& $-$
& $-$
& $-$
& $-$
& $-$
& $-$
& $-$
& \cellcolor{black!10}\underline{0.58}
& \cellcolor{black!10}0.60
& $-$
& $-$
\\

\hspace{15pt}TaggerOne
& 0.46
& $-$
& $-$
& $-$
& $-$
& $-$
& $-$
& \cellcolor{black!10}\textbf{0.82}
& \cellcolor{black!10}\textbf{0.83}
& $-$
& $-$
& $-$
& $-$
\\

\hspace{15pt}DNorm
& 0.22
& $-$
& $-$
& $-$
& $-$
& $-$
& $-$
& \cellcolor{black!10}\underline{0.73}
& $-$
& $-$
& $-$
& $-$
& $-$
\\

\hspace{15pt}KIRIs
& $-$
& $-$
& $-$
& $-$
& $-$
& $-$
& \cellcolor{black!10}\textbf{0.61}
& $-$
& $-$
& $-$
& $-$
& $-$
& $-$
\\

\hspace{15pt}CONORM
& $-$
& $-$
& $-$
& $-$
& 0.17
& \cellcolor{black!10}\textbf{0.53}
& $-$
& $-$
& $-$
& $-$
& $-$
& $-$
& $-$
\\


\bottomrule
\end{tabular*}

\caption{Exact-match Mention-Detection F1 Score. Bold marks the best measured score per dataset, underline the second best. Grey cells mark in-distribution results. \textit{Avg} represents the average score across the datasets, and \textit{Rank} is the average ranking per dataset for each model.}
\label{tab:ner-exact}
\end{table*}

\begin{table*}[t]
\centering
\normalsize
\setlength{\tabcolsep}{2.0pt}
\renewcommand{\arraystretch}{1.08}

\begin{tabular*}{\textwidth}{@{\extracolsep{\fill}}
    l
    *{11}{c}
    @{\hspace{4pt}}|@{\hspace{4pt}}
    c@{\hspace{1pt}}
    c
    @{}}

\toprule

& \multicolumn{9}{c}{MedPath}
& \multicolumn{2}{c}{Phenotype}
& \multicolumn{2}{c}{Overall} \\

\cmidrule(lr){2-10}
\cmidrule(lr){11-12}
\cmidrule(lr){13-14}

Model
& SC & MAN & MedM & COM
& ADR & CAD & MIM & NCBI
& CDR & BioC & GSC
& Avg & Rank \\

\midrule


\makecell[l]{\hspace{-4pt}\textit{LLM-based}}
& & & & & & & & & & & & & \\

\hspace{15pt}UMCR
& \textbf{0.54}
& \textbf{0.60}
& \textbf{0.40}
& \textbf{0.31}
& \textbf{0.62}
& 0.22
& \underline{0.30}
& 0.40
& \textbf{0.56}
& \underline{0.56}
& 0.58
& \textbf{0.46}
& \textbf{1.36}
\\

\hspace{15pt}AutoPCR
& 0.38
& \underline{0.44}
& \underline{0.32}
& 0.14
& 0.36
& 0.20
& 0.19
& 0.34
& 0.41
& \textbf{0.60}
& \underline{0.67}
& \underline{0.37}
& \underline{2.55}
\\

\hspace{15pt}REAL
& 0.24
& 0.39
& 0.09
& \underline{0.28}
& 0.09
& \underline{0.24}
& 0.17
& 0.21
& 0.23
& 0.43
& 0.32
& 0.24
& 4.09
\\

\midrule


\makecell[l]{\hspace{-4pt}\textit{Rule-based}}
& & & & & & & & & & & & & \\

\hspace{15pt}NCBO
& \underline{0.45}
& 0.24
& 0.00
& 0.11
& \underline{0.50}
& 0.18
& 0.15
& 0.16
& 0.25
& 0.40
& 0.55
& 0.27
& 4.36
\\

\hspace{15pt}QuickUMLS
& \underline{0.45}
& 0.20
& 0.06
& 0.10
& 0.26
& 0.17
& 0.14
& 0.33
& 0.27
& 0.38
& 0.41
& 0.25
& 4.64
\\

\hspace{15pt}FastHPOCR
& \underline{0.45}
& 0.22
& 0.17
& 0.07
& 0.34
& 0.07
& 0.10
& 0.27
& 0.35
& 0.48
& \textbf{0.72}
& 0.29
& 4.09
\\

\midrule


\makecell[l]{\hspace{-4pt}\textit{Supervised}}
& & & & & & & & & & & & & \\

\hspace{15pt}PhenoTagger
& $-$
& $-$
& $-$
& $-$
& $-$
& $-$
& $-$
& $-$
& $-$
& \cellcolor{black!10}0.54
& \cellcolor{black!10}0.57
& $-$
& $-$
\\

\hspace{15pt}TaggerOne
& 0.28
& $-$
& $-$
& $-$
& $-$
& $-$
& $-$
& \cellcolor{black!10}\textbf{0.63}
& \cellcolor{black!10}\underline{0.53}
& $-$
& $-$
& $-$
& $-$
\\

\hspace{15pt}DNorm
& 0.15
& $-$
& $-$
& $-$
& $-$
& $-$
& $-$
& \cellcolor{black!10}\underline{0.55}
& $-$
& $-$
& $-$
& $-$
& $-$
\\

\hspace{15pt}KIRIs
& $-$
& $-$
& $-$
& $-$
& $-$
& $-$
& \cellcolor{black!10}\textbf{0.43}
& $-$
& $-$
& $-$
& $-$
& $-$
& $-$
\\

\hspace{15pt}CONORM
& $-$
& $-$
& $-$
& $-$
& 0.15
& \cellcolor{black!10}\textbf{0.54}
& $-$
& $-$
& $-$
& $-$
& $-$
& $-$
& $-$

\\

\bottomrule
\end{tabular*}

\caption{Exact-match Linking F1 Score. Bold marks the best measured score per dataset, underline the second best. Grey cells mark in-distribution results. \textit{Avg} represents the average score across the datasets, and \textit{Rank} is the average ranking per dataset for each model.}
\label{tab:main-exact}
\end{table*}

\section{Prompts}
\label{app:prompts}
Figures~\ref{fig:prompt-s2}--\ref{fig:prompt-s4} show the brief-arm prompts for Stages 2--4. Fixed template text is verbatim; dataset-specific content from the compiled guideline brief and the input document is shown highlighted.

\begin{figure*}[t]
\centering
\begin{tikzpicture}[
    card/.style={
        rectangle, rounded corners=8pt, draw=black, very thick,
        inner sep=8pt, text width=0.94\textwidth, align=left,
        font=\sffamily\small
    }
]
\providecommand{\ph}[1]{\colorbox{black!12}{#1}}

\node[card] {
{\centering\sffamily\Large\bfseries Stage 2 --- Mention extraction\\[6pt]}
{\ttfamily\small
\# Guideline context  \\[4pt]
Stage 2 --- Mention extraction (high recall). \\
Find every explicit mention for the target entities, including \\
repeats, short forms, abbreviations as they appear in TEXT, \\
and signs/symptoms/abnormalities when they match the entity \\
definition. Return the \textbf{verbatim surface string(s)} copied \\
from the TEXT block (character offsets are computed later --- \\
do not invent numeric offsets). Prefer short precise surfaces \\
over long sentences. Prefer one contiguous surface per mention; \\
for true discontinuous mentions return an ordered list of \\
fragment strings. Do \textbf{not} assign concept IDs yet. Do \\
\textbf{not} drop mentions just because they might later fail \\
an exclusion or inventory rule --- later stages handle that. \\[4pt]
\#\#\# Guideline \\
\ph{COMPILED GUIDELINE} \\[4pt]
Focus entity: \ph{ENTITY\_ID} \\[4pt]
TEXT: \\
\ph{FULL DOCUMENT TEXT} \\[4pt]
Return a single JSON object: \\
\{ "mentions": [ \{ "text": "...", "confidence": 0.0 \} ] \} \\
Copy \texttt{text} EXACTLY as it appears in TEXT --- same casing, \\
spelling, whitespace, punctuation. No paraphrase, no offsets. \\
Discontinuous mention: \texttt{text} is a JSON list of fragments. \\
Annotate every occurrence. No concept IDs. If none: \\
\{ "mentions": [] \}.
}};
\end{tikzpicture}
\caption{Stage 2 mention-extraction prompt.}
\label{fig:prompt-s2}
\end{figure*}

\begin{figure*}[t]
\centering
\begin{tikzpicture}[
    card/.style={
        rectangle, rounded corners=8pt, draw=black, very thick,
        inner sep=8pt, text width=0.94\textwidth, align=left,
        font=\sffamily\small
    }
]
\providecommand{\ph}[1]{\colorbox{black!12}{#1}}

\node[card] {
{\centering\sffamily\Large\bfseries Stage 3 --- Agentic linking\\[6pt]}
{\ttfamily\small
\# Guideline context \\[4pt]
Stage 3 --- Concept linking (specificity-seeking). \\
For the focus mention, use ontology tools to search and \\
inspect parents/children/siblings, then commit the most \\
specific eligible concept ID supported by the mention and \\
document context (or unlink). You may slightly adjust the \\
mention surface if the span boundary is wrong, without \\
inventing a new mention. Prefer a single ID unless the \\
brief says otherwise. \\[4pt]
\#\#\# Guideline \\
\ph{COMPILED GUIDELINE} \\[4pt]
Focus mention: \ph{MENTION SURFACE} \\
TEXT: \\
\ph{FULL DOCUMENT TEXT} \\
Sibling mentions: \ph{OTHER MENTIONS IN DOC} \\[4pt]
Respond with a single JSON object, ONE action per turn: \\[2pt]
search --- ranked candidate CUIs (names + scores) \\
\ \ \{"action": "search", "query": "...", "k": 10\} \\
get\_concept --- name, definition, synonyms, types \\
\ \ \{"action": "get\_concept", "id": "..."\} \\
get\_parents --- broader / less-specific concepts \\
get\_children --- narrower / more-specific concepts \\
get\_siblings --- same-level alternatives \\
adjust\_span --- refine boundaries (expand/shrink only): \\
\ \ \{"action": "adjust\_span", "text": "..."\} \\
Terminal: \{"action": "commit", "cui": "..."\} or \\
\ \ \{"action": "unlink", "reason": "..."\} \\
Before commit, explore the graph around the current \\
favorite concept; switch to a better-supported parent or \\
child and repeat. Commit only when no other concept is \\
more adequate. Max 8 steps; exhausted loops unlink.
}};
\end{tikzpicture}

\begin{tikzpicture}[
    card/.style={
        rectangle, rounded corners=8pt, draw=black, very thick,
        inner sep=8pt, text width=0.94\textwidth, align=left,
        font=\sffamily\small
    }
]
\providecommand{\ph}[1]{\colorbox{black!12}{#1}}

\node[card] {
{\centering\sffamily\Large\bfseries Stage 3 --- Graph-exploration variants\\[6pt]}
{\ttfamily\small
\textbf{Optional.} Goal sentence reads ``... use ontology tools \\
to search and \textbf{optionally} inspect parents/children/ \\
siblings ...''. No graph-exploration instruction. \\[4pt]
\textbf{Encouraged.} Goal sentence reads ``... use ontology \\
tools to search and inspect parents/children/siblings ...''. \\
No graph-exploration instruction. \\[4pt]
\textbf{Mandatory (default).} Encouraged goal sentence, plus: \\
``Before commit, explore the graph around the current \\
favorite concept. If a child or parent is better supported \\
by the mention and local context, switch to it and repeat. \\
Commit only when no other concept is more adequate.''
}};
\end{tikzpicture}
\caption{Stage 3 agentic-linking prompt. Lower panel: graph-exploration variants from the agentic ablation (Optional, Encouraged, Mandatory=default).}
\label{fig:prompt-s3}
\end{figure*}

\begin{figure*}[t]
\centering
\begin{tikzpicture}[
    card/.style={
        rectangle, rounded corners=8pt, draw=black, very thick,
        inner sep=8pt, text width=0.94\textwidth, align=left,
        font=\sffamily\small
    }
]
\providecommand{\ph}[1]{\colorbox{black!12}{#1}}

\node[card] {
{\centering\sffamily\Large\bfseries Stage 4 --- Guideline filter \\[6pt]}
{\ttfamily\small
\# Guideline context \\[4pt]
Stage 4 --- Guideline conformance filter. \\
Given mention--concept pairs and the guideline, decide keep / \\
unlink / drop for soft exclusion and edge-case rules. Every \\
rejection must cite a stable rule\_id and a short reason. \\
Hard inventory filters may already have been applied. \\[4pt]
\#\#\# Guideline \\
\ph{COMPILED GUIDELINE} \\[4pt]
Pairs: \\
\ph{MENTION--CUI PAIRS FOR THIS DOCUMENT} \\[4pt]
Return a single JSON object: \\
\{ "decisions": [ \{ "mention\_id": "...", \\
\ \ "action": "keep" / "unlink" / "drop", \\
\ \ "rule\_id": "...", "reason": "..." \} ] \} \\
Decide EVERY listed mention exactly once. Prefer unlink \\
when the span may stay but the CUI is invalid; drop when \\
the mention should not be annotated at all. rule\_id MUST \\
be one of the Enforceable rules --- never invent one. \\
Reason $\le$12 words, no commentary outside JSON. If unsure \\
and the pair looks guideline-legal, keep.
}};
\end{tikzpicture}
\caption{Stage 4 guideline-filter prompt (brief arm).}
\label{fig:prompt-s4}
\end{figure*}

\section{Guidelines}
\label{app:guideline_compilation}

\begin{figure*}[t]
\centering
\input{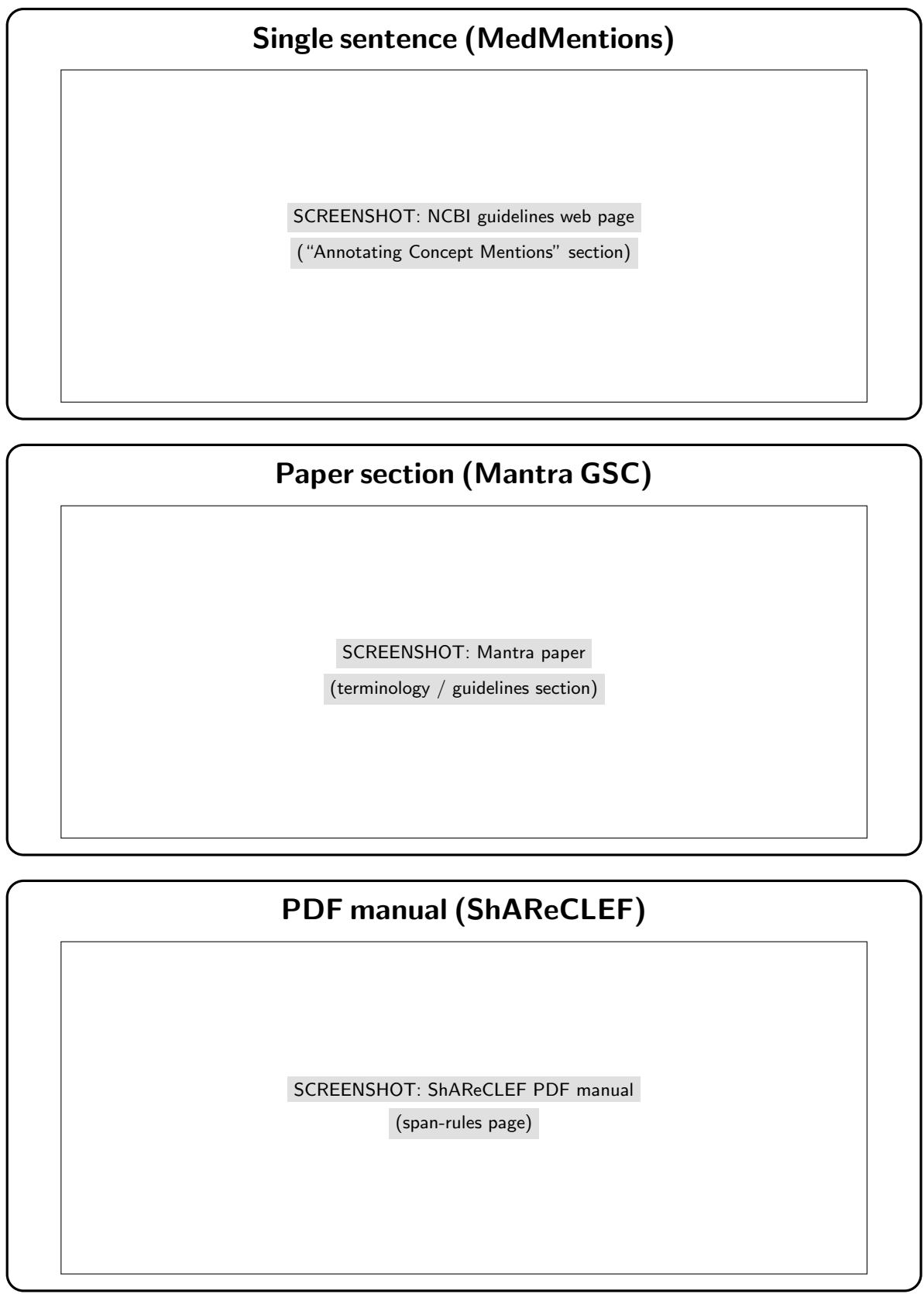}
\caption{Guideline sources: paper sentence (MedMentions), a paper section (Mantra GSC), and a full PDF manual (ShAReCLEF).}
\label{fig:guideline-examples}
\end{figure*}

\paragraph{Guideline Examples.}
Figure~\ref{fig:guideline-examples} shows guideline sources at three scales, from paper sentences to a full PDF manual. Each panel holds a screenshot of the source.

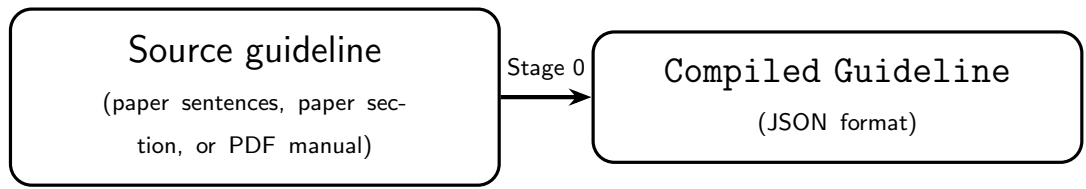
\begin{figure*}[t]
\centering
\begin{tikzpicture}[
    card/.style={
        rectangle, rounded corners=8pt, draw=black, very thick,
        inner sep=10pt, text width=0.36\textwidth, align=center,
        font=\sffamily\Large
    }
]
\providecommand{\ph}[1]{\colorbox{black!12}{#1}}

\node[card] (src) {Source guideline \\[2pt] {\small (paper sentences, paper section, or PDF manual)}};
\node[card, right=1.2cm of src] (out) {\texttt{Compiled Guideline} \\[2pt] {\small (JSON format)}};
\draw[-{Stealth[length=3mm]}, black, ultra thick] (src) --
    node[midway, above=2pt, font=\sffamily\small] {Stage 0} (out);
\end{tikzpicture}
\caption{Guideline compilation: source guideline to compiled JSON (Stage 0).}
\label{fig:guideline-compilation}
\end{figure*}

\paragraph{Guideline Compilation.}
Figure~\ref{fig:guideline-json} gives the compiled \texttt{GuidelineBrief} JSON format, Figure~\ref{fig:guideline-compilation} shows the Stage 0 compilation step conceptually, and Figure~\ref{fig:guideline-bare} shows the bare-guideline counterpart with no compiled content.

\begin{figure*}[t]
\centering
\begin{tikzpicture}[
    card/.style={
        rectangle, rounded corners=8pt, draw=black, very thick,
        inner sep=8pt, text width=0.94\textwidth, align=left,
        font=\sffamily\small
    }
]
\providecommand{\ph}[1]{\colorbox{black!12}{#1}}

\node[card] {
{\centering\sffamily\Large\bfseries Compiled \texttt{GuidelineBrief} JSON format\\[6pt]}
{\ttfamily\small
\{ \\
\ \ "schema\_version": "2.0.0", \\
\ \ "dataset\_id": \ph{DATASET ID}, \\
\ \ "dataset\_name": \ph{DATASET NAME}, \\
\ \ "task\_summary": \ph{TASK SUMMARY (1--3 SENTENCES)}, \\
\ \ "entities": [ \{ "entity\_id": \ph{ID}, "name": \ph{NAME}, \\
\ \ \ \ "definition": \ph{DEFINITION} \} ], \\
\ \ "ontology": \ph{ONTOLOGY SCOPE / SUBSET CONSTRAINTS}, \\
\ \ "include": [ \ph{INCLUDE BULLETS} ], \\
\ \ "exclude": [ \ph{EXCLUDE BULLETS} ], \\
\ \ "spans": \ph{SPAN POLICY}, \\
\ \ "linking": \ph{LINKING POLICY}, \\
\ \ "notes": [ \ph{NOTES} ], \\
\ \ "rules": [ \{ "id": \ph{RULE ID}, "stage": \ph{STAGE}, \\
\ \ \ \ "target": \ph{SPAN / CUI / BOTH}, "text": \ph{RULE TEXT} \} ] \\
\}
}};
\end{tikzpicture}
\caption{\texttt{Compiled Guideline} JSON format (placeholders; no dataset content).}
\label{fig:guideline-json}
\end{figure*}

\begin{figure*}[t]
\centering
\begin{tikzpicture}[
    card/.style={
        rectangle, rounded corners=8pt, draw=black, very thick,
        inner sep=8pt, text width=0.94\textwidth, align=left,
        font=\sffamily\small
    }
]
\providecommand{\ph}[1]{\colorbox{black!12}{#1}}

\node[card] {
{\centering\sffamily\Large\bfseries Bare guideline\\[6pt]}
{\ttfamily\small
Stage 2 --- Mention extraction (high recall). \\
Find every explicit mention for the target entities \\
\# Target entity : \ph{[placeholder]}
}};
\end{tikzpicture}
\caption{Bare guideline: stage goals only, no compiled brief.}
\label{fig:guideline-bare}
\end{figure*}
\end{document}